\documentclass[10pt,twocolumn,letterpaper]{article}

\usepackage{cvpr}              

\definecolor{cvprblue}{rgb}{0.21,0.49,0.74}
\usepackage[pagebackref,breaklinks,colorlinks,allcolors=cvprblue]{hyperref}
\usepackage{multirow} 
\usepackage[accsupp]{axessibility}  
\usepackage{makecell}
\usepackage{float}
\usepackage{graphicx}
\usepackage{subcaption} 

\def\paperID{13245} 
\def\confName{CVPR}
\def\confYear{2025}

\title{End-to-End Implicit Neural Representations for Classification }

\author{Alexander Gielisse, Jan van Gemert\\
Delft University of Technology\\
}

\begin{document}
\maketitle
\begin{abstract}

Implicit neural representations (INRs)  such as NeRF and SIREN encode a signal in neural network parameters and show excellent results for signal reconstruction. Using INRs for downstream tasks, such as classification, is however not straightforward. Inherent symmetries in the parameters pose challenges and current works primarily focus on designing architectures that are equivariant to these symmetries. However, INR-based classification still significantly under-performs compared to pixel-based methods like CNNs. This work presents an end-to-end strategy for initializing SIRENs together with a learned learning-rate scheme, to yield representations that improve classification accuracy. We show that a simple, straightforward, Transformer model applied to a meta-learned SIREN, without incorporating explicit symmetry equivariances, outperforms the current state-of-the-art. On the CIFAR-10 SIREN classification task, we improve the state-of-the-art without augmentations from 38.8\% to 59.6\%, and from 63.4\% to 64.7\% with augmentations. We demonstrate scalability on the high-resolution Imagenette dataset achieving reasonable reconstruction quality with a classification accuracy of 60.8\% and are the first to do INR classification on the full ImageNet-1K dataset where we achieve a SIREN classification performance of 23.6\%. To the best of our knowledge, no other SIREN classification approach has managed to set a classification baseline for any high-resolution image dataset. Our code is available at \url{https://github.com/SanderGielisse/MWT}.

\end{abstract}    
\section{Introduction}
\label{sec:intro}

Implicit neural representations (INRs)~\cite{nerf, wire, gaussian_act, data_to_functa, spatial_functa, spatial_acorn, spatial_instant_ngp, spatial_liif, spatial_mdif, spatial_miner, spatial_neurbf} encode complex continuous signals compactly in a neural network parameterized by $\theta$. In the case of images, an INR continuously maps spatial image coordinates \(x, y \in \mathbb{R}\) to RGB values. Specifically, we define the function \( f_{\theta}: \mathbb{R}^2 \rightarrow \mathbb{R}^3 \) such that:
\[
f_{\theta}(x, y) = (r, g, b),
\]
where \((x, y)\) denotes continuous spatial coordinates, \(r, g, b \in \mathbb{R}\) are the RGB values at that position. By doing so, \(\theta\) encodes the image implicitly, instead of explicitly in a pixel value grid.

A common model choice for $f_{\theta}$ is a multilayer perceptrons (MLP) \cite{siren, nerf, wire, gaussian_act}, which enable high-quality reconstructions. We identify two main advantages of using MLP-based implicit neural representations (INRs). First, the capacity of the model \( f_{\theta}(x, y) \) is not necessarily uniformly distributed across the image space, unlike representations based on fixed-resolution pixel grids. Second, the signal used as input is not required to be an equidistant pixel grid; any subset of observations from a signal can be used to train \( f_{\theta}(x, y) \). Unfortunately, while INRs are highly effective for high-resolution reconstructions, using these implicit representations directly for downstream tasks, such as classification, remains challenging, as it requires reasoning on the parameters $\theta$.

\begin{figure*}[h]
    \centering
    \includegraphics[width=0.515\linewidth]{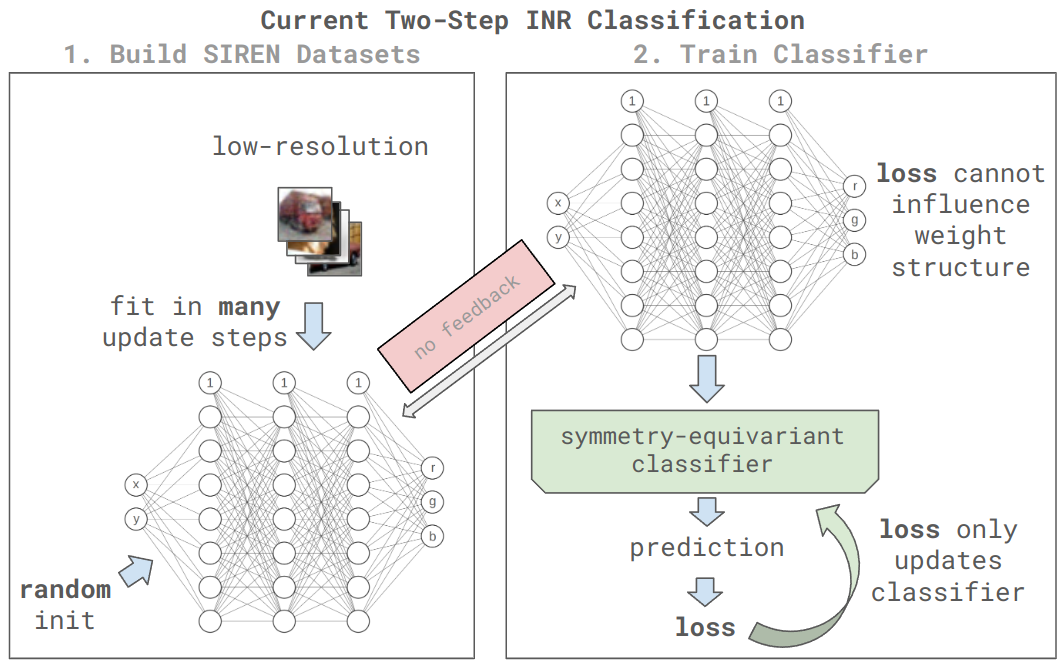}
    \hspace{0.3cm}
    \includegraphics[width=0.45\linewidth]{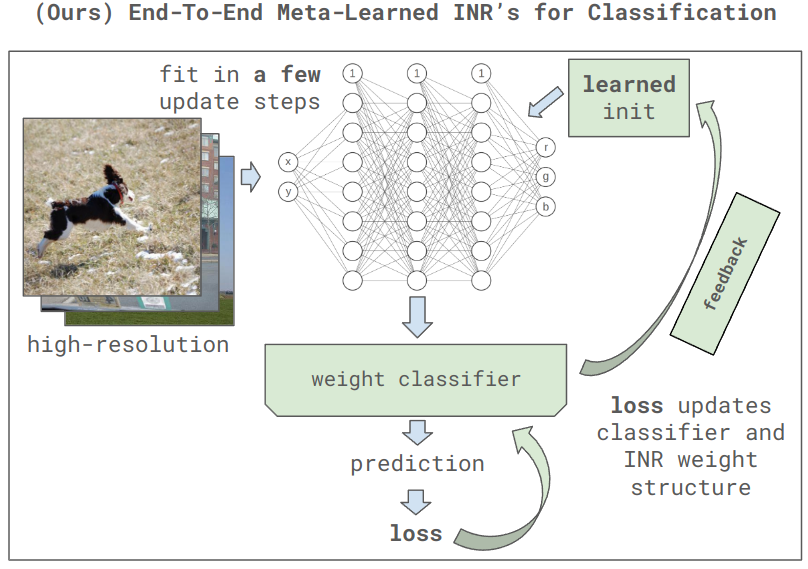}
    \caption{\textbf{Left:} the existing two-step INR classification approach, where the process involves building INR datasets by fitting images using many update steps without classifier feedback for its weight structure. The classifier is trained separately, and the classification loss cannot influence the INR. \textbf{Right:} the proposed end-to-end meta-learned INR approach for classification. This method fits high-resolution images using a few update steps with learned initialization and feedback from the classifier, allowing the classification loss to update both the classifier and the INR weight structure, enhancing downstream performance while ensuring quick convergence.}
    \label{fig:overview}
\end{figure*}

To perform a downstream task such as classification on the parameters \(\theta\), an additional model \( g \) is required, which takes \(\theta\) as input. This involves constructing a model architecture that can process the weights of another architecture as its input. However, \(\theta\) can contain many symmetries. For example, in the case of MLPs, reordering the nodes and their associated weights introduces permutation symmetries; that is, a different arrangement of weights that corresponds to the exact same function. Similarly, scale-symmetries allows scaling parameters in a way that results in an identical function, even though \(\theta\) has changed.

One approach to address these symmetries is to realign the weights so that all symmetries map to the same network. Unfortunately, this alignment problem is intractable~\cite{deep_weight_space_alignment, git_rebasin}. An alternative solution is to design the downstream architecture \( g \) to be equivariant to the symmetries in \( \theta \), effectively bypassing the alignment issue. Consequently, many recent works have adopted this equivariant design approach for the design of the downstream architecture~\cite{inr2vec, perm_eq_neural_functionals, graph_equivariant_representations, graph_processing_diverse, scale_equivariant_graph_metanetworks, dws_net}. However, the performance of these approaches remains behind that of pixel-based classification methods. One possible reason for this could be that RGB pixel-based representations are inherently easier for a downstream model to interpret than the weights of another neural network. It may be that the weights $\theta$ of the INR lack sufficient ``structure'', making it difficult for downstream models to identify useful image features. The cause of sub-par performance being the absence of structure is supported by the findings of \cite{fit_a_nef}, who find that using the same, shared, INR initialization for all images and then make image-specific INRs by updating the shared initialization for each particular image INR yields improved classification results. This shared initialization may avoid symmetries by picking a fixed reference point. This is corroborated by the observations of \cite{fit_a_nef} that more INR update steps to minimize the reconstruction loss improves the reconstruction quality, it actually reduces classification performance. Essentially, as the weights diverge from the initial shared INR, the structure becomes increasingly distinct, making classification of the INR more challenging. For this reason, we balance the INR fitting process between reconstruction quality and classification performance, in an end-to-end meta-learning setting using only a few update steps. What is more, using only a few update steps allows for a highly efficient implementation, which allows INR classification on high-resolution images and fitting INRs on data-augmentations in the interpretable image space.

Rather than merely sharing a common INR initialization across all images, we enforce structural consistency beyond shared initialization. Namely, we propose a meta-learning approach to jointly optimize the INR initialization and the learning rate schedule used in the image-specific updates. The objective is to find an initialization and a learning rate schedule such that, when an INR is fitted to an image signal, the resulting parameters \(\theta\) have a structure that is directly interpretable for the classification model \(g(\theta)\). Rather than the current two-step approach of first converting a set of images into their INR representations and as an independent next step training a classifier, we make the INR fitting process part of the classifier training loop. We do this by back-propagating through the INR optimization steps themselves. This way, an end-to-end trainable setting is achieved where the classifier can influence the structure of the INR, see Figure \ref{fig:overview}. We focus specifically on classification as the downstream task and use the popular SIREN as our MLP-based INR as commonly done \cite{inr2vec, perm_eq_neural_functionals, scale_equivariant_graph_metanetworks, dws_net}, although our approach is likely applicable to other tasks and other INRs as well. Our contributions are as follows.

\begin{itemize}

\item \textbf{End-to-end learning of INR classification}: Development of a meta-learning initialization strategy for SIRENs, combined with a meta-learned learning-rate scheme, aimed at enhancing classification performance on SIRENs. 

\item \textbf{Computational efficiency}: The high convergence speed of our approach allows high-resolution image classification and enables the use of interpretable spatial image augmentations during training. We explore a computationally efficient variant where a SIREN is learned on a subset of the pixels in each step. This further reduces computational cost, without deterioration of reconstruction quality or classification accuracy.

\item \textbf{Simple classifier design}: We apply a straightforward, standard,  Transformer model to the meta-learned SIREN representations, achieving improvements in classification accuracy without requiring complex classifiers explicitly designed to be equivariant to weight symmetries. 

\item \textbf{State of the art classification results}: We improve the SOTA on MNIST \cite{mnist} accuracy from $96.6\%$ to $98.8\%$, Fashion-MNIST \cite{fashionmnist} from $80.8\%$ to $90.4\%$ and CIFAR-10 \cite{cifar10} from $38.8\%$ to $59.6\%$. For CIFAR-10, we also use augmentations and improve the SOTA from $63.4\%$ (inr2array \cite{neural_functional_transformers}) to $64.7\%$. Moreover, to our knowledge, we are the first to set SIREN classification baselines on high-resolution datasets. On Imagenette \cite{imagenette} we achieve $60.8\%$ classification accuracy, while on the full ImageNet-1K we achieve $23.63\%$ accuracy.

\item \textbf{Comprehensive ablation study}: Detailed ablation studies on key components of the proposed meta-learning and Transformer-based approach, analyzing the impact of meta-initialization, learning-rate schemes, and Transformer architecture choices on reconstruction and classification performance.

\end{itemize}

\section{Related Work}
\label{sec:related}

\textbf{Implicit Neural Representations.}
INRs can be defined as a single differentiable and continuous function over the entire signal space, globally, often parameterized by an MLP~\cite{siren, wire, gaussian_act, data_to_functa}. This allows flexible INRs capacity allocation, using less parameters on simpler areas. A key drawback, however, is that these models can take a long time to converge and are challenging to train to a low reconstruction error. An alternative is a hybrid explicit-implicit variant, where multiple local implicit representations are explicitly arranged spatially, either uniformly across the image or based on specific heuristics \cite{spatial_acorn, luijmes2025arcanchoredrepresentationclouds, spatial_functa, spatial_instant_ngp, spatial_liif, spatial_mdif, spatial_miner, spatial_neurbf}. In our work, we  do not consider hybrid local models and focus only on fully implicit, global, models.

For global MLP-based INRs, ReLU activations~\cite{gaussian_act} struggle to represent high-frequency components, often requiring positional embeddings~\cite{fourier_features}. The seminal SIREN \cite{siren} approach, on the other hand, demonstrates that a specific initialization with \(\sin\) activation functions achieves smooth convergence, high reconstruction quality, and well-defined higher-order derivatives without requiring positional embeddings leading to several follow-up works~\cite{gaussian_act,wire}. In our work, we adopt the SIREN approach due to its widespread adoption~\cite{use_siren_pigan, scale_equivariant_graph_metanetworks, data_to_functa, spatial_functa}, straightforward implementation, and proven effectiveness in preserving high-frequency details across diverse images.

\smallskip
\textbf{MLP-based INR Classification.} Classifying INRs is challenging as it requires reasoning on the MLP weights of the INR. The work by INSP-Net \cite{inspnet} addresses this by differential operators, directly applied to INRs, enabling both low-level image-to-image translation and more complex tasks like classification. Alternatively, INR2VEC \cite{inr2vec} uses a feature encoder for each node in the MLP, followed by max-pooling to obtain a global feature vector. However, this approach does not account for potential symmetries present in the neuron weights.

\smallskip
\textbf{Equivariant INR Classifiers.} DWS-Net \cite{dws_net} proposes a network that can take as input an INR parameterized by an MLP, by processing it through a series of layers that are equivariant to the permutation symmetries of an MLP. NFN \cite{perm_eq_neural_functionals} improves upon this by making stronger symmetry assumptions to improve parameter efficiency and practical scalability. Graph Metanetworks \cite{graph_processing_diverse, graph_equivariant_representations} extend this work to not only accept MLPs as input, but show that any set of operations that can be described by a graph.

Neural Functional Transformers (NFTs) \cite{neural_functional_transformers} define an attention-based architecture to process the weights of MLPs and CNNs, and propose their INR2ARRAY method to convert an INR into its permutation invariant latent representation. Moreover, while most work has been focused on equivariance to the inherent permutation symmetries, recent work has also explored the use of graph meta-networks that are scale-equivariant \cite{scale_equivariant_graph_metanetworks} to the scale symmetries in the weights. However, in our work, we will take a different approach. Namely, we demonstrate that enforcing structure on the MLP parameters offers an alternative to the downstream symmetry equivariance approach, showing that strong performance can be achieved without explicitly modeling equivariances in the classifier, allowing the use of straightforward, standard, classifiers.

\begin{figure}
    \centering
    \includegraphics[width=\linewidth]{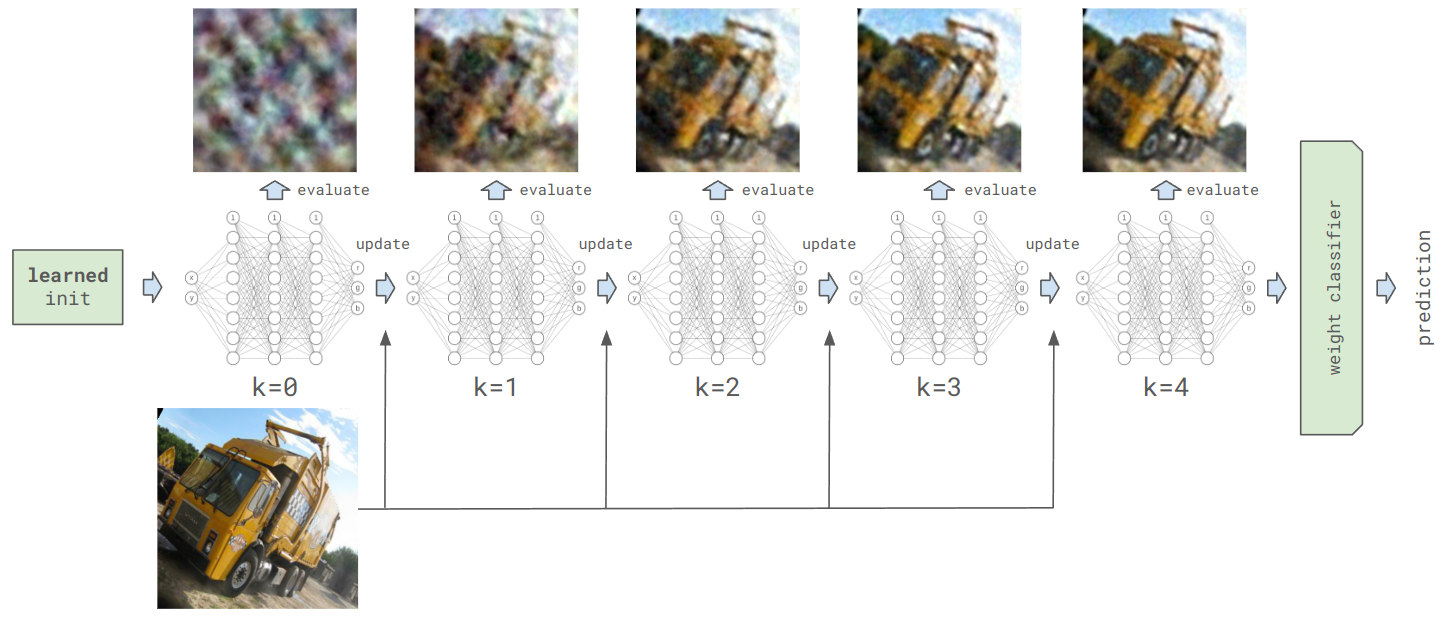}
    \caption{ Our method illustrated for a high-resolution Imagenette \cite{imagenette} image. A meta-learned SIREN initialization  is updated for a small amount of gradient steps, in this case $k=4$. The resulting weights are then passed to a  classifier. Meta-learning allows us to back-propagate through the update steps, end-to-end optimizing the SIREN both for reconstruction as for classification. }
    \label{fig:inference_overview}
\end{figure}

\smallskip
\textbf{Differentiable Meta-Learning.} Model-Agnostic Meta-Learning (MAML) \cite{maml} is a meta-learning technique, with the objective of finding a set of model initialization parameters $\theta$ that can be quickly adapted, in $k$ 'inner loop' steps, using only a few samples. Meta-SGD \cite{meta_sgd} extends MAML by also learning the learning rates for the $k$ steps. This approach uses the property that the gradient descent operator itself is differentiable,  \ie,
\begin{equation}
    \theta_{k+1} = \theta_{k} - \alpha \nabla_\theta \mathcal{L}_{\text{inner}}( \cdot ; \theta_{k}),
\end{equation}
is differentiable with respect to the learning rate $\alpha$ and to $\theta_{k}$ if the loss function $\mathcal{L}_{\text{inner}}$ is also differentiable with respect to $\theta_k$.

Then, for this $\theta_k$, we can compute a different task loss $\mathcal{L}_{\text{task}}$. This task loss can be minimized by back-propagating through the gradient steps of the inner-loop to update the initial $\theta_0$, such that after $k$ steps of minimizing $\mathcal{L}_{\text{inner}}$ we obtain $\theta_k$ that minimizes $\mathcal{L}_{\text{task}}$. While prior work typically uses the same loss function for both $\mathcal{L}_{\text{inner}}$ and $\mathcal{L}_{\text{task}}$, we demonstrate that explicitly incorporating a classification loss into $\mathcal{L}_{\text{task}}$ leads to improved classification performance. We use both MAML and Meta-SGD for meta-learning the INR initialization and learning rate schedule, to balance reconstruction as well as downstream classification. 

\smallskip
\textbf{Meta-Learning for INRs.} Approaches similar to MAML are used in \cite{meta_nerf} to fit an implicit 3D NeRF~\cite{nerf} in just a few update steps. The data-to-functa~\cite{data_to_functa} approach uses a meta-learning to speed up the INR fitting process while also doing dimensionality reduction with reconstruction quality as the only optimization goal, although follow-up work shows that this does not scale to high-resolution images~\cite{spatial_functa}. In fit-a-nef~\cite{fit_a_nef} they manually tune and investigate meta properties and confirm that sharing an initialization and using only a few update steps helps classification. Our work differs in that we do not want to manually tune such parameters, but use meta-learning to automatically learn a balance between good reconstruction and classification accuracy, using the benefits of a shared initialization with the computational advantages of having only a few update steps, for high-resolution images. To our knowledge, no previous works use classification performance as an optimization goal in the INR meta-learning.

\section{Method: Meta Weight Transformer (MWT)}
\label{sec:method}

\newcommand{\gradZero}{
g_{\theta, \alpha}^{\text{rec}} \leftarrow \nabla_{\theta, \alpha}\mathcal{L}_{\text{rec}}(f_\phi, \bf{x})
}
\newcommand{\gradOne}{
g_{\theta, \alpha}^{\text{cls}} \leftarrow  \nabla_{\theta, \alpha}\mathcal{L}_{\text{cls}}(\hat{y}, y)
}
\newcommand{\gradTwo}{
g_{\psi} \leftarrow \nabla_{\psi}\mathcal{L}_{\text{cls}}(\hat{y}, y)
}

\newcommand{\gradThree}{
g_{\theta, \alpha} \leftarrow g_{\theta, \alpha}^{\text{rec}} + g_{\theta, \alpha}^{\text{cls}} * w_{\text{cls}} 
}

\subsection{Meta-Learned INR}
We use SIREN \cite{siren} as the INR, and follow their initialization scheme to generate the very first random initial $\theta$. We meta-learn this SIREN initialization $\theta$ so it can be used as a shared starting point for all images for fitting an image-specific SIREN, which is updated $k$ steps for each individual image. We also learn a learning rate schedule $\alpha \in \mathbb{R}^{k \times |\theta|}$ which contains a learning rate for each parameter at each of the $k$ update steps, based on \cite{meta_sgd}. The objective when learning this $\theta$ and $\alpha$ is not only to ensure rapid convergence in only $k$ steps in terms of good reconstruction quality, but also to obtain SIREN parameters that are directly interpretable by a simple downstream INR classifier.

\begin{algorithm}
	\caption{
            Task-Specific SIREN Meta-Learning.\\
            \small{
            For the inner-loop, we minimize a reconstruction loss $\mathcal{L}_{\text{rec}}$. We then optimize the initial SIREN parameters $\theta$ and learning rate schedule $\alpha$ such that $\phi$ does not only encode the image with high quality, but is also in a format that can be correctly classified by our classifier $h_{\psi}(\phi)$. Here, $f$ is our SIREN, $w_{\text{cls}}$ is a scalar that we can use to change the classifier influence on the meta-learning process, and Adam refers to the use of the Adam \cite{adam} optimizer.
            }
    }\label{alg:metalearn}

	\begin{algorithmic}[1]
 
            \State Init random SIREN with parameters $\theta$
            \State Init learning rates $\alpha$ for all $k$ update steps $\alpha \in \mathbb{R}^{k \times |\theta|}$
            \While {not converged} \hfill \texttt{<outer loop>}
                \State Sample training image $\bf{x}$ with classification label $y$ 
                \State Set starting INR parameters to shared base $\phi = \theta$
                \For {$i=(1,2,\ldots, k)$} \hfill \texttt{<inner loop>}
                    \State $\phi \leftarrow \phi - \alpha_{i} \nabla_{\phi}\mathcal{L}_{\text{rec}}(f_\phi, \bf{x})$
                \EndFor
                \State Predict classification label $\hat{y} \leftarrow h_{\psi}(\phi)$
                \State Get $\theta, \alpha$ gradient $\gradOne$
                \State Get $\theta, \alpha$ gradient   $\gradZero$
                \State Combine $\theta, \alpha$ gradients $\gradThree$
                \State Update SIREN initialization $\theta \leftarrow \text{Adam}(\theta, g_{\theta})$
                \State Update learning rates $\alpha \leftarrow \text{Adam}(\alpha, g_{\alpha})$ 
                \State Get $\psi$ gradient $\gradTwo$
                \State Update classifier $\psi \leftarrow \text{Adam}(\psi, g_{\psi})$
            \EndWhile
	\end{algorithmic} 
\end{algorithm}

For clarity, we explain our method using a batch size of $1$ with a single data point of one image $\bf x$ with class label $y$. In Figure~\ref{fig:inference_overview} we show a visual overview.  We explain the method below, and pseudo-code is given in Algorithm~\ref{alg:metalearn}.

The meta-learning makes use of two loops: an outer loop over all training images, and an inner loop to update the shared learned initialization $\theta$ which is updated $k$ times for each specific image $\bf x$ in an inner loop. We denote with $\phi$ the SIREN parameters $\theta$ after updating them $k$ times. So, we define a downstream model $h_{\psi}(\phi)$ that takes as input the $k$ times updated sample-specific SIREN parameters $\phi$. To update the initial $\theta$ to become $\phi$, we set $\mathcal{L}_{\text{inner}}$ to be the MSE reconstruction loss of the SIREN reconstruction  $f_\theta(\cdot)$ for each of the $k$ inner-loop steps, for all $n$ pixels in image $\bf x$ as

\begin{equation}
 \mathcal{L}_{\text{rec}}(f_\phi, {\bf x}) = \frac{1}{n} \sum_{j=1}^n ({\bf x}_j - f_\phi(\mathbf{x}_j) )^2
\end{equation}

We cannot use the classification loss in the inner-loop, as it has to be performed at test time for new samples, where we do not have the ground truth $y$ of that sample. After the inner-loop, we classify $\phi$ to the predicted class label $\hat{y}$ using our INR classifier $\hat{y} \leftarrow h_{\psi}(\phi)$. Then, we compute both the reconstruction loss $\mathcal{L}_{\text{rec}}(f_\phi, \bf{x})$ and the classifier loss $\mathcal{L}_{\text{cls}}(\hat{y}, y)$. First, we update the parameters of the meta-learning process. For both these losses, we compute the gradients with respect to both the initialization values $\theta$, and learning rates $\alpha$.  We denote these gradients as $g_{\theta, \alpha}^{\text{rec}}$ and $g_{\theta, \alpha}^{\text{cls}}$ which gives

\begin{equation}\gradZero \end{equation}
\begin{equation}\gradOne \end{equation}

Then, we combine the gradients using a scalar $w_{\text{cls}}$, with which we can control the influence of the classifier on the meta-learning process, to obtain $g_{\theta, \alpha}$

\begin{equation}\gradThree \end{equation}

which we use to obtain the gradients that we will use to update $\theta$ and $\alpha$. Lastly, we compute the gradients to update the classifier weights $\psi$. We denote this gradient as $g_{\psi}$ which we compute as

\begin{equation}\gradTwo \end{equation}

As the weights of the classifier $\psi$ are solely optimized for classification performance, we only use $g_{\psi}$ to update them.

\subsection{SIREN Classification using Transformers}

To classify our image-specific SIREN parameters $\phi$, we require an architecture $h_{\psi}(\phi)$ that is capable of accepting linear layers with bias terms as input. We 
use a simple, straight-forward, Transformer model \cite{attention_all_you_need} that operates directly on the weight space of a SIREN model without considering any explicit INR symmetries.

To input the SIREN network to the classifier we use the property that a linear transformation followed by the addition of biases can be combined into a single matrix operation, if we pad each layer with a $1$-valued input. Specifically, the weights of a linear layer \(W_L \in \mathbb{R}^{c_{\text{in}} \times c_{\text{out}}}\) and the bias term \(W_b \in \mathbb{R}^{c_{\text{out}}}\) can be represented together as \(W \in \mathbb{R}^{(c_{\text{in}}+1) \times c_{\text{out}}}\). This combined matrix can be applied to the input vector padded with a 1, resulting in an operation that is identical to applying the weight matrix followed by the bias addition. We show a SIREN network this way in Figure \ref{fig:overview}.

We train the classifier  on the hidden layers of the SIREN. SIREN networks often maintain the same layer dimensionality throughout their depth. This means that, if we only consider the hidden layers, each neuron in a SIREN network has \(c_{in} + 1\) input weights. We use each output neuron of the SIREN's hidden layers to be a token for our Transformer model, where each neuron has a \(c_{\text{in}} + 1\) dimensional feature vector; those being the input weights to that neuron.

So, for a SIREN network with \(l\) hidden layers and dimensionality \(d\), there are \(d \times l\) input tokens for our Transformer model. The feature vectors for these tokens do not inherently have any positional bias, making it challenging for the Transformer to identify which neuron a specific feature vector corresponds to. To address this, we introduce a learned positional bias $\beta \in \mathbb{R}^{|\theta|}$ for each weight. Additionally, we observe that the SIREN base weights \(\theta\) and the image-specific weights after the inner loop \(\phi\) are often similar. Namely, element-wise $|\theta-\phi|$ results in mostly small numbers. Consequently, our classifier struggles to interpret these weights accurately, likely due to the low-frequency bias of our model \cite{fourier_features}. To mitigate this, we provide the network with the difference between the base SIREN weights $\theta$ and the image-specific weights $\phi$ instead, rather than the absolute values.  Because this difference often results in small values, and the commonly used Transformer initializations expect normalized inputs, we scale the feature vectors by a scalar \(\lambda\), which we typically set to 500. So, before providing the weights to the Transformer, we first convert our SIREN as 
\begin{equation}
    \phi_{\text{scaled}} \leftarrow \lambda(\phi - \theta + \beta). 
\end{equation}
\paragraph{Reducing Computational Cost} Computationally, the reconstruction loss in the inner loop is typically calculated for each pixel at each of the $k$ inner-loop steps, requiring $H \times W \times k$ forward passes through the SIREN network, where $H$ and $W$ represent the image height and width in pixels, respectively. The computation graph for all $k$ steps and for all $H \times W$ pixels must be stored to facilitate meta-learning of the base initialization $\theta$, which can become resource-intensive for high-resolution images. To address this, we also explore an alternative approach that fits the SIREN to the image using a random subset $S$ of pixels at each inner-loop step, rather than processing all pixels. We denote the fraction of pixels used with $s \in [0, 1]$ such that $|S| = (s \times H \times W$). Note that when $s$ is set to $(1 / k)$, on average the model still sees every pixel once. By decreasing $s$, we effectively make the $k$ inner-loop steps more stochastic, while simultaneously saving on computational cost.



\begin{table}[htbp]
\centering
\resizebox{\columnwidth}{!}{
\begin{tabular}{p{3cm} ccc}
\toprule

& \multicolumn{3}{c}{Classification Accuracy (\%)}\\
\textbf{Method} & \textbf{MNIST} & \textbf{Fashion-MNIST} & \textbf{CIFAR-10} \\
\midrule
MLP & 17.55 $\pm$ 0.01* & 19.91 $\pm$ 0.47* & 11.38 $\pm$ 0.34* \\
Inr2Vec \cite{inr2vec} & 23.69 $\pm$ 0.10* & 22.33 $\pm$ 0.41* & - \\
DWS \cite{dws_net} & 85.71 $\pm$ 0.57 & 67.06 $\pm$ 0.29 & 34.45 $\pm$ 0.42* \\
NFN$_{\text{NP}}$ \cite{perm_eq_neural_functionals} & 78.50 $\pm$ 0.23* & 68.19 $\pm$ 0.28* & 33.41 $\pm$ 0.01* \\
NFN$_{\text{HNP}}$ \cite{perm_eq_neural_functionals} & 79.11 $\pm$ 0.84* & 68.94 $\pm$ 0.64* & 28.64 $\pm$ 0.07* \\
NG-GNN \cite{graph_equivariant_representations} & 91.40 $\pm$ 0.60 & 68.00 $\pm$ 0.20 & 36.04 $\pm$ 0.44* \\
ScaleGMN \cite{scale_equivariant_graph_metanetworks} & 96.57 $\pm$ 0.10 & 80.46 $\pm$ 0.32 & 36.43 $\pm$ 0.41 \\
ScaleGMN-B  \cite{scale_equivariant_graph_metanetworks} & 96.59 $\pm$ 0.24 & 80.78 $\pm$ 0.16 & 38.82 $\pm$ 0.10 \\
\midrule

WT (Ours) &  91.38  $\pm$ 1.67 &  83.97  $\pm$ 1.38 &  43.78  $\pm$ 0.64 \\
MWT (Ours) &  98.33  $\pm$ 0.11 &  89.41  $\pm$ 0.25 &  56.90  $\pm$ 0.29 \\
MWT-L (Ours) &  \textbf{98.80  $\pm$ 0.06} &  \textbf{90.43  $\pm$ 0.23} &  \textbf{59.57  $\pm$ 0.52} \\

\bottomrule
\end{tabular}
}
\caption{Accuracy of various methods on MNIST, Fashion-MNIST, and CIFAR-10 datasets where models operate directly on the SIREN weights. The values represent the mean $\pm$ standard deviation ($n=3$), in the no data-augmentation setting. The star * denotes that the numbers were taken from \cite{scale_equivariant_graph_metanetworks}, the (Ours) rows are computed by us, and other values are taken from the original papers. We show that end-to-end meta-learned classifier gradients (MWT) strongly improves over not having such classifier gradients (WT). Moreover, MWT, as well as our large model MWT-L set a strong new baseline for single SIREN classification.}

\label{table:performance_comparison}
\end{table}

\section{Experiments}
\label{sec:experiments}

\textbf{Implementation details.} 
The classifier is a 10 block Transformer~\cite{attention_all_you_need}, each block with a dimensionality matching that of the SIREN network. LayerNorm \cite{layernorm} is applied prior to the self-attention operation and the fully connected GELU \cite{gelu} layer. We implement multi-head attention with a head dimension of 64 and use LayerScale \cite{layerscale} initialized at $0.1$. For optimization, we employ the AdamW \cite{adam, adamw} optimizer with a batch size of 16, a learning rate of \(1 \times 10^{-4}\), and a weight decay of \(1 \times 10^{-4}\). 

In the inner-loop optimization described in Algorithm \ref{alg:metalearn}, we use plain SGD without momentum. We observed that a higher AdamW learning rate of \(1 \times 10^{-2}\) for the inner-loop learning rates $\alpha$ improves performance, so we adopt this value there. The inner-loop learning rates for meta-SGD are initialized as \(\alpha \sim \text{Uniform}(0.1, 1.0)\). We set the rescaling factor for SIREN weights to \(\lambda = 500\). For the SIREN network itself, we use a \(\omega = 10.0\) on the first layer, a hidden dimensionality of 128, and a depth of 4, unless specified otherwise. All experiments were conducted on NVIDIA A40 GPUs with mixed-precision enabled.

\subsection{Classification of INRs}

We evaluate INR classification on MNIST \cite{mnist}, Fashion-MNIST \cite{fashionmnist}, and CIFAR-10 \cite{cifar10}. We train each model on each of the datasets for 10 epochs. 

\vspace{-3mm}

\paragraph{Impact of End-to-End Classifier Gradients}

We investigate our end-to-end Meta Weight Transformer (MWT) to verify if gradients from the classifier effectively influence the meta-learned SIREN initialization and learning rate schedule. Thus, we compare MWT to a variant of Algorithm~\ref{alg:metalearn} that does not back-propagate the $\mathcal{L}_{\text{cls}}$ loss through the inner loop and instead uses it solely for optimizing the downstream Transformer model. We label this variant as the Weight Transformer (WT), for which we set $w_{\text{cls}} = 0$. In this modified version, the downstream Transformer classifier does not influence the meta-learned initialization nor the learning rate schedule, leading to a learning procedure that is learned purely for fast and high-quality reconstruction. This WT approach resembles the traditional two-step method outlined in Figure \ref{fig:overview}. Here, the INR is trained separately with only a reconstruction objective, after which a classifier is trained on the INRs separately.

The results in Table~\ref{table:performance_comparison} show a large improvement of MWT over WT. This indicates that incorporating end-to-end classifier gradients to guide the structuring of the INR through meta-learning is advantageous. 

\paragraph{Comparisons to State-of-the-art}

The Table~\ref{table:performance_comparison} also includes a comparison with current state-of-the-art results in SIREN classification where no augmentations are used. Our WT model mostly performs similar or better than previous state-of-the-art models, with MWT clearly outperforming all earlier methods. 

A possible explanation for why WT already does well, is that our SIREN is learned through the meta-learning process designed to achieve convergence within just a few update steps. Namely, while non-meta learning approaches typically require several dozen update steps to achieve reasonable reconstruction accuracy, our method uses only \( k = 6 \) update steps. This reduction in the number of steps may contribute to the observed improvement in classification performance, consistent with findings from \cite{fit_a_nef}, which demonstrated that reducing the number of update steps can enhance downstream classification performance. 

Comparing to other INR baselines that do use augmentations on CIFAR-10, these report $41.27\%$~\cite{dws_net}, $46.60\%$~\cite{perm_eq_neural_functionals}, $56.95\%$~\cite{scale_equivariant_graph_metanetworks}, and $63.4$\%~\cite{neural_functional_transformers}. Our MWT-L models scores 59.57\% $\pm$ 0.52, which is quite reasonable with respect to computation effort, because the augmented models typically need to fit 10x-50x more SIRENS. Furthermore, we scale our MWT approach further, use a 20-layer Transformer, use spatial augmentations, increase epochs to 40, and achieve a new CIFAR-10 \cite{cifar10} SIREN classification SOTA of 64.7\%.

\subsection{Upscaling to High-Resolution}

So far we focused on low-resolution images, we now explore the scalability of our  approach to high-resolution images. To begin, we train our MWT classifier on Imagenette \cite{imagenette}, a 10-class subset of the ImageNet dataset \cite{imagenet}, with augmentations enabled, as we clarify below.

\paragraph{Spatial Augmentations} When constructing datasets composed of SIREN networks, image-space augmentations such as scaling, rotation, flipping, and translation can be computationally expensive, because it requires re-training the SIREN to each new augmented signal. To address this, weight-space augmentations have been proposed \cite{dws_net, weight_aug}. However, augmenting the weight space in a way such that it actually still represents the type of images that can naturally occur is challenging. Fortunately, since we can fit a SIREN to a new signal in just a few gradient steps, we can simply apply image-space augmentations. Therefore, in the following experiment, we also evaluate model variants where spatial image augmentations of the training data are used. We decrease the number of inner-loop steps to $k=4$, as this is computationally less costly, and the results in Table \ref{table:ablation} did not show a large drop in performance for this change. Furthermore, we investigate if we can push the performance by using our larger model variant MWT-L with SIREN dimensionality of $256$. By increasing the SIREN dimensionality, we also automatically increase the dimensionality of the Transformer tokens to $256$. To facilitate the higher image resolution, we use $\omega = 30.0$ for the first layer of the SIREN for all of the high-resolution models. We vary the amount of sub-sampling by setting different values for $s$. We present the results in Table \ref{tab:performance_imagenette}. Note that for $s=0.25$ with $k=4$, the model on average sees every pixel once, similar to traditional image processing models.

First, we compare the WT and MWT models. Surprisingly, we find that the best performance for classification using the small models is achieved for MWT with augmentations while only using 5\% of all the pixels in each of the reconstruction inner-loop steps, resulting in a validation accuracy of $57.27\%$ on the high-resolution Imagenette \cite{imagenette} dataset. Interestingly, using smaller values for $s$ does not appear to strongly impact classification performance nor reconstruction quality. A similar pattern is found for our large MWT-L model, where the classification accuracy, as well as the reconstruction quality, is relatively consistent between varying $s$. This suggests that the meta-learned SIREN network may have learned an implicit image bias that enables it to interpolate the missing pixels with sufficient accuracy, perhaps similar to \cite{deep_image_prior}, or simply that just a partial signal observation is sufficient to achieve the current performance. This might allow the model to maintain performance even when provided with incomplete data during training, suggesting potential further applications in scenarios involving sparse input data \cite{sparse_conv} or point cloud processing.

\paragraph{ImageNet-1K} We train MWT-L on the full large-scale ImageNet-1K \cite{imagenet} dataset. We use a dimensionality of $256$, use spatial image augmentations, use a subsampling rate of $0.01$ for each of the $k=4$ inner-loop steps. In Table \ref{tab:imagenet} we show the performance for varying $w_{\text{task}}$. To our knowledge, our work is the first to train SIREN networks for a high-resolution collection of images such as ImageNet \cite{imagenet}.

\begin{table*}
\centering
\resizebox{1.9\columnwidth}{!}{%
\begin{tabular}{p{2.2cm} ccccccccc}
\toprule

\textbf{Model} & \multicolumn{2}{c}{\textbf{Accuracy (\%)}} & \multicolumn{2}{c}{\textbf{PSNR (dB)}} & \textbf{\#Params} & \textbf{\#Params} &
\textbf{Validation} & \textbf{Training} & \textbf{Memory} \\
    & No Aug. &  With Aug. & No Aug. &  With Aug. & CLS & SIREN & (sec / epoch) & (min / epoch) & (GiB) \\
\midrule

$\text{WT}_{s=0.25}$ &  47.96  $\pm$ 0.29 &  46.83  $\pm$ 0.25 &  \textbf{23.23  $\pm$ 0.08} &  \textbf{23.27  $\pm$ 0.05} &  1.1M &  332K &  40.7 &  5.0 &  19.6 \\
$\text{WT}_{s=0.1}$ &  48.11  $\pm$ 0.42 &  47.02  $\pm$ 0.37 &  22.50  $\pm$ 0.02 &  22.47  $\pm$ 0.04 &  1.1M &  332K &  20.5 &  2.3 &  9.0 \\
$\text{WT}_{s=0.05}$ &  49.01  $\pm$ 0.17 &  47.97  $\pm$ 1.27 &  21.64  $\pm$ 0.03 &  21.79  $\pm$ 0.04 &  1.1M &  332K &  12.7 &  1.4 &  5.6 \\
\midrule
$\text{MWT}_{s=0.25}$ &  48.79  $\pm$ 0.36 &  56.13  $\pm$ 0.28 &  21.14  $\pm$ 0.02 &  21.14  $\pm$ 0.08 &  1.1M &  332K &  41.4 &  8.0 &  21.1 \\
$\text{MWT}_{s=0.1}$ &  50.14  $\pm$ 0.87 &  56.78  $\pm$ 0.27 &  20.82  $\pm$ 0.04 &  21.14  $\pm$ 0.02 &  1.1M &  332K &  20.5 &  3.5 &  9.6 \\
$\text{MWT}_{s=0.05}$ &  50.23  $\pm$ 0.31 &  57.27  $\pm$ 0.38 &  20.68  $\pm$ 0.03 &  20.98  $\pm$ 0.06 &  1.1M &  332K &  13.5 &  2.1 &  5.9 \\
\midrule
$\text{MWT-L}_{s=0.1}$ &  51.80  $\pm$ 0.23 &  60.62  $\pm$ 0.37 &  21.94  $\pm$ 0.02 &  22.31  $\pm$ 0.03 &  4.3M &  1.3M &  42.4 &  7.6 &  24.1 \\
$\text{MWT-L}_{s=0.05}$ &  52.04  $\pm$ 0.45 &  60.75  $\pm$ 0.65 &  21.56  $\pm$ 0.02 &  21.86  $\pm$ 0.04 &  4.3M &  1.3M &  28.7 &  4.7 &  17.4 \\
$\text{MWT-L}_{s=0.02}$ &  \textbf{52.93  $\pm$ 0.90} &  \textbf{60.75  $\pm$ 0.37} &  20.95  $\pm$ 0.04 &  21.14  $\pm$ 0.02 &  4.3M &  1.3M &  21.2 &  2.9 &  13.5 \\

\bottomrule
\end{tabular}%
}
\caption{Accuracy and PSNR results for WT, MWT and MWT-L models with varying subsampling settings on the Imagenette \cite{imagenette} dataset. The columns (with aug.) indicate the use of spatial augmentations to the training data; scaling, translating, rotating and flipping.
Note that due to random uniform subsampling, also the PSNR is computed on a random, though \textit{different}, uniform subset of the pixels. So, the shown PSNR is just an approximation of the real PSNR.}
\label{tab:performance_imagenette}
\end{table*}

\begin{table}[htbp]
\centering
\resizebox{0.6\columnwidth}{!}{
\begin{tabular}{p{2.2cm} ccc}
\toprule

\textbf{Model} & \textbf{Accuracy} & \textbf{PSNR} \\
\midrule
$w_{\text{task}}=0.01$ &  \textbf{24.11\%} &  21.78 dB \\
$w_{\text{task}}=0.001$ &  21.87\% &  22.09 dB \\
$w_{\text{task}}=0.0001$ &  19.21\% & \textbf{22.14} dB \\

\bottomrule
\end{tabular}
}
\caption{Results for the full ImageNet-1K \cite{imagenet} dataset trained for 40 epochs. Accuracy indicates the percentage of correctly classified samples, while PSNR represents the average reconstruction quality of the image. This model uses 12.1 GiB of memory for training, takes $306.0 \pm 4.2$ minutes for a single epoch, takes $216.8 \pm 9.4$ seconds for making predictions on the full validation set. The SIREN model has 1.3M parameters, and the WT classifier has 4.5M parameters.}
\label{tab:imagenet}
\end{table}

\begin{figure}[h]
    \centering
    \begin{subfigure}[h]{0.6\columnwidth}
        \centering
        \includegraphics[width=\columnwidth]{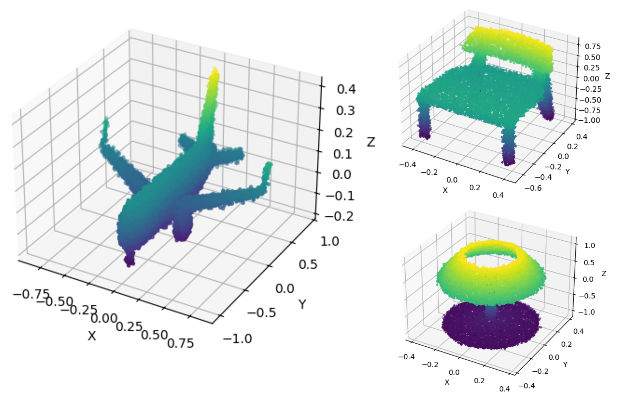} 
    \end{subfigure}
    
    \hfill
    \begin{subfigure}[h]{\columnwidth}
        \centering
        \resizebox{\columnwidth}{!}{ 
            \begin{tabular}{@{}lcccc@{}}
            \toprule
            \textbf{MWT-L Model} \\
            \textbf{Subsampling} & \textbf{Accuracy} & \textbf{Validation} & \textbf{Training Time} & \textbf{Memory} \\
            (parameter $s$) & (\%) & (sec / epoch) & (min / epoch) & (train, GiB) \\
            \midrule

            0.1 &  85.82 &  116.2 &  7.8 &  15.2 \\
            0.01 &  85.41 &  49.5 &  3.7 &  11.0 \\
            0.001 &  84.04 &  46.7 &  3.5 &  10.8 \\
            0.00005 &  30.35 & 41.7  &  3.1 &  10.7 \\
            
            \bottomrule
            \end{tabular}
        }
    \end{subfigure}
    \caption{ModelNet40 \cite{modelnet40} results on unsigned distance functions. Trained for 150 epochs using dimensionality of $256$, number of inner-loop steps $k=4$, with augmentations enabled. Timings include the inner-loop fitting of the INR. The inr2vec work \cite{inr2vec} scores an accuracy of $87.0\%$ on this dataset, with non-INR approaches like PointNet \cite{pointnet} and PointNet++ \cite{pointnet_pp} outperforming these with accuracies of $88.8\%$ and $89.7\%$ respectively.}
\end{figure}

\subsection{Ablations}

To analyze the impact of our model's hyperparameters, we conduct an ablation study using the CIFAR-10 dataset, with results presented in Table \ref{table:ablation}. Notably, when examining the task influence on the meta-learning of the SIREN initialization, denoted as \( w_{\text{cls}} \), we observe a clear trade-off between reconstruction quality and classification performance. Specifically, if \( w_{\text{cls}} \) in Algorithm~\ref{alg:metalearn} is set too high, both reconstruction quality and classification accuracy decline. On the other hand, if \( w_{\text{cls}} \) is set too low, only classification performance is negatively affected. We find that setting \( w_{\text{cls}} = 0.01 \) strikes a good balance, yielding good classification performance while maintaining acceptable reconstruction quality. For the number of inner-loop steps \( k \), we observe that taking more gradient steps generally enhances both classification and reconstruction performance. However, higher values of \( k \) can be computationally expensive, which is a known restriction of MAML \cite{maml}, so we use \( k=6 \) as a practical compromise. Similarly, increasing the depth of the Transformer and the width of the SIREN network also improves performance but leads to a substantial increase in the number of model parameters. We set the SIREN dimensionality to $128$ for our MWT model, and to $256$ for our MWT-L model.

\begin{table}[htbp]
\centering
\resizebox{\columnwidth}{!}{
\begin{tabular}{p{3cm} cccccccc}
\toprule

\textbf{Ablation} & \textbf{Parameter} & \textbf{Accuracy} & \textbf{PSNR} & \textbf{Validation} & \textbf{Training} & \textbf{Mem} & \textbf{\#Params} & \textbf{\#Params} \\
    & & (\%) & (dB) & (sec/epoch) & (min/epoch) & (GiB) & CLS & SIREN \\
\midrule

\multirow{6}{*}{\textbf{Task Influence} $w_{\text{task}}$}
 & 0 & 42.24 & 	43.11 & 	14.0 & 	3.1 & 	3.0 & 	1.1M & 	465K \\
 & 0.001 & 53.53 & 	38.92 & 	14.2 & 	4.0 & 	3.0 & 	1.1M & 	465K \\
 & \underline{0.01} & 55.79 & 	30.93 & 	15.2 & 	4.3 & 	3.0 & 	1.1M & 	465K \\
 & 0.1 & 54.00 & 	23.11 & 	15.5 & 	4.2 & 	3.0 & 	1.1M & 	465K \\
 & 1.0 & 52.35 & 	20.89 & 	14.2 & 	4.2 & 	3.0 & 	1.1M & 	465K \\
 & 10.0 & 52.34 & 	18.14 & 	14.0 & 	4.1 & 	3.0 & 	1.1M & 	465K \\
\midrule
\multirow{6}{*}{\textbf{Inner-Loop Steps $k$}}
 & 1 & 49.70 & 	23.63 & 	6.7 & 	1.7 & 	2.2 & 	1.1M & 	133K \\
 & 2 & 52.46 & 	25.24 & 	9.1 & 	2.5 & 	2.3 & 	1.1M & 	199K \\
 & 4 & 54.79 & 	29.10 & 	11.0 & 	3.1 & 	2.6 & 	1.1M & 	332K \\
 & \underline{6} & 55.79 & 	30.93 & 	15.2 & 	4.3 & 	3.0 & 	1.1M & 	465K \\
 & 8 & 55.52 & 	30.96 & 	16.9 & 	5.0 & 	3.3 & 	1.1M & 	598K \\
 & 10 & 55.58 & 	30.79 & 	19.8 & 	6.0 & 	3.6 & 	1.1M & 	731K \\
\midrule
\multirow{6}{*}{\textbf{Scaling Factor $\lambda$}}
 & 1 & 48.60 & 	31.38 & 	15.0 & 	4.2 & 	3.0 & 	1.1M & 	465K \\
 & 10 & 54.80 & 	31.71 & 	13.9 & 	4.0 & 	3.0 & 	1.1M & 	465K \\
 & 100 & 55.16 & 	31.43 & 	14.0 & 	4.1 & 	3.0 & 	1.1M & 	465K \\
 & \underline{500} & 55.79 & 	30.93 & 	15.2 & 	4.3 & 	3.0 & 	1.1M & 	465K \\
 & 1000 & 55.54 & 	30.59 & 	14.0 & 	4.1 & 	3.0 & 	1.1M & 	465K \\
 & LayerNorm & 54.89 & 	31.18 & 	14.0 & 	4.1 & 	3.0 & 	1.1M & 	465K \\
\midrule
\multirow{4}{*}{\textbf{SIREN Depth}}
 & 2 & 55.39 & 	28.50 & 	10.5 & 	2.9 & 	1.5 & 	1.0M & 	234K \\
 & \underline{4} & 55.79 & 	30.93 & 	15.2 & 	4.3 & 	3.0 & 	1.1M & 	465K \\
 & 6 & 55.52 & 	31.96 & 	21.9 & 	6.2 & 	4.9 & 	1.1M & 	696K \\
 & 8 & 55.79 & 	32.73 & 	31.6 & 	8.5 & 	7.4 & 	1.1M & 	927K \\
\midrule
\multirow{3}{*}{\textbf{SIREN Width}}
 & 64 & 51.54 & 	27.65 & 	9.0 & 	3.3 & 	1.1 & 	273K & 	118K \\
 & \underline{128} & 55.79 & 	30.93 & 	15.2 & 	4.3 & 	3.0 & 	1.1M & 	465K \\
 & 256 & 57.19 & 	33.15 & 	43.6 & 	11.2 & 	12.7 & 	4.3M & 	1.8M \\
\midrule
\multirow{4}{*}{\textbf{Transformer Depth}}
 & 5 & 54.41 & 	31.03 & 	12.2 & 	3.6 & 	2.2 & 	581K & 	465K \\
 & \underline{10} & 55.79 & 	30.93 & 	15.2 & 	4.3 & 	3.0 & 	1.1M & 	465K \\
 & 15 & 56.77 & 	30.99 & 	15.8 & 	4.6 & 	3.7 & 	1.6M & 	465K \\
 & 20 & 56.82 & 	30.82 & 	18.0 & 	5.2 & 	4.5 & 	2.1M & 	465K \\
\midrule
\multirow{2}{*}{\textbf{Meta-SGD Shared}}
 & \underline{False} & 55.79 & 	30.93 & 	15.2 & 	4.3 & 	3.0 & 	1.1M & 	465K \\
 & True & 54.38 & 	29.86 & 	15.2 & 	4.2 & 	3.0 & 	1.1M & 	133K \\
\midrule
\multirow{3}{*}{\textbf{Meta-SGD LR}}
 & \underline{0.01} & 55.79 & 	30.93 & 	15.2 & 	4.3 & 	3.0 & 	1.1M & 	465K \\
 & 0.001 & 53.34 & 	27.82 & 	13.9 & 	4.1 & 	3.0 & 	1.1M & 	465K \\
 & 0.0001 & 50.58 & 	25.95 & 	15.2 & 	4.2 & 	3.0 & 	1.1M & 	465K \\
\midrule
\multirow{4}{*}{\textbf{Subsampling Rate $s$}}
 & \underline{1.0} & 55.79 & 	30.93 & 	15.2 & 	4.3 & 	3.0 & 	1.1M & 	465K \\
 & 0.1 & 54.21 & 	24.06 & 	11.5 & 	3.4 & 	2.1 & 	1.1M & 	465K \\
 & 0.01 & 46.45 & 	18.84 & 	12.0 & 	3.5 & 	2.0 & 	1.1M & 	465K \\
 & 0.001 & 18.03 & 	16.53 & 	12.0 & 	3.4 & 	2.0 & 	1.1M & 	465K \\

\bottomrule
\end{tabular}
}
\caption{Full ablation results with timings and memory usage for the proposed MWT, trained on 80\% of the CIFAR-10 training dataset. The reported numbers are evaluated on the remaining 20\% of the training set that was held out for validation. Accuracy indicates the percentage of correctly classified samples, while PSNR represents the average reconstruction quality of the image.}
\label{table:ablation}
\end{table}

\begin{figure}[H]
    \centering
    \includegraphics[width=0.8\linewidth]{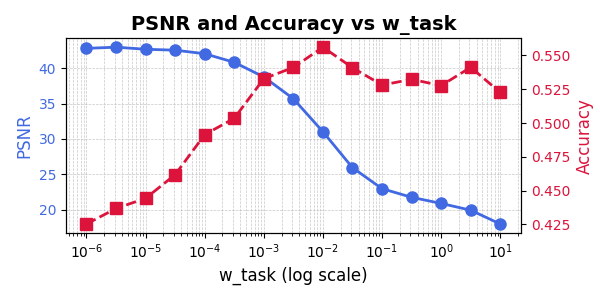}
    \label{fig:your-label}
    \vspace{-0.4cm}
    \caption{We provide a visualization of the trade-off made in the ablation of the MWT model. Increasing the influence of the classifier on the meta-learning of the INR ($w_{\text{task}}$) decreases reconstruction quality, but increases classification performance up to about $0.01$, after which both decrease.}
\end{figure}

\newcommand{\nfnnp}{$\text{NFN}_{\text{NP}}$ }
\newcommand{\nfnhnp}{$\text{NFN}_{\text{HNP}}$ }

\begin{table}[H]
\centering
\resizebox{0.9\columnwidth}{!}{%
\begin{tabular}{@{}lcccccccc@{}}
\toprule
\textbf{Classifier} & \textbf{Classifier Influence} & \textbf{Accuracy} & \textbf{PSNR} &
\textbf{\#Params} & \textbf{\#Params} & \textbf{Validation} & \textbf{Training} & \textbf{Memory} \\
& & (\%) & (dB) & CLS & SIREN & (test set, sec) & (min / epoch) & (GiB) \\
\midrule

\multirow{2}{*}{\textbf{WT}}
 & ($w_{\text{task}} = 0$) &  38.98 &  \textbf{43.09} &  1.1M &  465K &  30.3 &  10.8 &  4.2 \\
 & ($w_{\text{task}} = 0.01$) &  \textbf{56.02} &  32.64 &  1.1M &  465K &  29.0 &  11.6 &  4.2 \\
\midrule

\multirow{3}{*}{\textbf{\nfnnp}}
 & ($w_{\text{task}} = 0$) &  29.54 &  \textbf{44.21} &  1.8M &  465K &  35.4 &  16.0 &  0.7 \\
 & ($w_{\text{task}} = 0.01$) &  \textbf{49.89} &  32.55 &  1.8M &  465K &  28.1 &  11.7 &  4.5 \\
\midrule

\multirow{3}{*}{\textbf{\nfnhnp}}
 & ($w_{\text{task}} = 0$) &  25.77 &  \textbf{44.41} &  3.6M &  465K &  35.6 &  14.2 &  5.2 \\
 & ($w_{\text{task}} = 0.01$) &  \textbf{48.37} &  32.83 &  3.6M &  465K &  28.0 &  11.7 &  5.8 \\

\bottomrule
\end{tabular}%
}
\caption{We show our method on the NFN classifier \cite{perm_eq_neural_functionals} on CIFAR-10 \cite{cifar10}. For NFN, we use three-layered network with a layer dimensionality of 64, followed by two fully connected layers. We train for 10 epochs, with spatial augmentations enabled. To compare, NFN~\cite{perm_eq_neural_functionals} reports $44.1\%$ for \nfnhnp, and an accuracy of $46.6$\% for \nfnnp, but do this through augmentation by fitting 20 SIRENs from different initializations for each image.}
\label{tab:subsampling}
\end{table}

\section{Discussion}
\label{sec:discussion}

We introduce a straightforward Transformer architecture for classifying SIREN networks, without being equivariant to any parameter symmetries of the SIREN. We integrate the SIREN fitting process into the classifier’s training loop in an end-to-end manner, allowing the classifier to adjust the initial parameterization and learning rate schedule of the SIREN for new images. This approach enables the classifier to shape the parameter space in a way that enhances classification performance, while also optimizing reconstruction quality. Our method establishes a new state-of-the-art baseline across multiple SIREN image classification benchmarks.

However, we are uncertain whether the initial parameterization and learning rate schedule alone can impose sufficient structure. Future work could explore additional learnable constraints on the SIREN parameters to further enhance classification performance. While we consider the current reconstruction quality sufficient for our work, improvements may be possible, especially for high-resolution datasets, which could be an area for further investigation.

We view this work as a step away from the conventional focus on achieving optimal reconstruction quality in implicit neural representations (INRs), shifting toward developing INRs that also optimize for classification accuracy. Future research on balancing the INR optimization between classification accuracy and reconstruction quality could further advance this direction. In principle, we do believe that continuous signals are best represented by a continuous function, instead of the accidental, discrete, sensor grid and as such we, explicitly, believe that INRs have a bright future.

{
    \small
    \bibliographystyle{ieeenat_fullname}
    \bibliography{main}

\begin{thebibliography}{46}
\providecommand{\natexlab}[1]{#1}
\providecommand{\url}[1]{\texttt{#1}}
\expandafter\ifx\csname urlstyle\endcsname\relax
  \providecommand{\doi}[1]{doi: #1}\else
  \providecommand{\doi}{doi: \begingroup \urlstyle{rm}\Url}\fi

\bibitem[Ainsworth et~al.(2022)Ainsworth, Hayase, and Srinivasa]{git_rebasin}
Samuel~K Ainsworth, Jonathan Hayase, and Siddhartha Srinivasa.
\newblock Git re-basin: Merging models modulo permutation symmetries.
\newblock \emph{arXiv preprint arXiv:2209.04836}, 2022.

\bibitem[Ba(2016)]{layernorm}
Jimmy~Lei Ba.
\newblock Layer normalization.
\newblock \emph{arXiv preprint arXiv:1607.06450}, 2016.

\bibitem[Bauer et~al.(2023)Bauer, Dupont, Brock, Rosenbaum, Schwarz, and Kim]{spatial_functa}
Matthias Bauer, Emilien Dupont, Andy Brock, Dan Rosenbaum, Jonathan~Richard Schwarz, and Hyunjik Kim.
\newblock Spatial functa: Scaling functa to imagenet classification and generation.
\newblock \emph{arXiv preprint arXiv:2302.03130}, 2023.

\bibitem[Chan et~al.(2021)Chan, Monteiro, Kellnhofer, Wu, and Wetzstein]{use_siren_pigan}
Eric~R Chan, Marco Monteiro, Petr Kellnhofer, Jiajun Wu, and Gordon Wetzstein.
\newblock pi-gan: Periodic implicit generative adversarial networks for 3d-aware image synthesis.
\newblock In \emph{Proceedings of the IEEE/CVF conference on computer vision and pattern recognition}, pages 5799--5809, 2021.

\bibitem[Chen et~al.(2021{\natexlab{a}})Chen, Liu, and Wang]{spatial_liif}
Yinbo Chen, Sifei Liu, and Xiaolong Wang.
\newblock Learning continuous image representation with local implicit image function.
\newblock In \emph{Proceedings of the IEEE/CVF conference on computer vision and pattern recognition}, pages 8628--8638, 2021{\natexlab{a}}.

\bibitem[Chen et~al.(2021{\natexlab{b}})Chen, Zhang, Genova, Fanello, Bouaziz, H{\"a}ne, Du, Keskin, Funkhouser, and Tang]{spatial_mdif}
Zhang Chen, Yinda Zhang, Kyle Genova, Sean Fanello, Sofien Bouaziz, Christian H{\"a}ne, Ruofei Du, Cem Keskin, Thomas Funkhouser, and Danhang Tang.
\newblock Multiresolution deep implicit functions for 3d shape representation.
\newblock In \emph{Proceedings of the IEEE/CVF International Conference on Computer Vision}, pages 13087--13096, 2021{\natexlab{b}}.

\bibitem[Chen et~al.(2023)Chen, Li, Song, Chen, Yu, Yuan, and Xu]{spatial_neurbf}
Zhang Chen, Zhong Li, Liangchen Song, Lele Chen, Jingyi Yu, Junsong Yuan, and Yi Xu.
\newblock Neurbf: A neural fields representation with adaptive radial basis functions.
\newblock In \emph{Proceedings of the IEEE/CVF International Conference on Computer Vision}, pages 4182--4194, 2023.

\bibitem[De~Luigi et~al.(2023)De~Luigi, Cardace, Spezialetti, Ramirez, Salti, and Di~Stefano]{inr2vec}
Luca De~Luigi, Adriano Cardace, Riccardo Spezialetti, Pierluigi~Zama Ramirez, Samuele Salti, and Luigi Di~Stefano.
\newblock Deep learning on implicit neural representations of shapes.
\newblock \emph{arXiv preprint arXiv:2302.05438}, 2023.

\bibitem[Deng et~al.(2009)Deng, Dong, Socher, Li, Li, and Fei-Fei]{imagenet}
Jia Deng, Wei Dong, Richard Socher, Li-Jia Li, Kai Li, and Li Fei-Fei.
\newblock Imagenet: A large-scale hierarchical image database.
\newblock In \emph{2009 IEEE conference on computer vision and pattern recognition}, pages 248--255. Ieee, 2009.

\bibitem[Dupont et~al.(2022)Dupont, Kim, Eslami, Rezende, and Rosenbaum]{data_to_functa}
Emilien Dupont, Hyunjik Kim, SM Eslami, Danilo Rezende, and Dan Rosenbaum.
\newblock From data to functa: Your data point is a function and you can treat it like one.
\newblock \emph{arXiv preprint arXiv:2201.12204}, 2022.

\bibitem[Finn et~al.(2017)Finn, Abbeel, and Levine]{maml}
Chelsea Finn, Pieter Abbeel, and Sergey Levine.
\newblock Model-agnostic meta-learning for fast adaptation of deep networks.
\newblock In \emph{International conference on machine learning}, pages 1126--1135. PMLR, 2017.

\bibitem[Graham and Van~der Maaten(2017)]{sparse_conv}
Benjamin Graham and Laurens Van~der Maaten.
\newblock Submanifold sparse convolutional networks.
\newblock \emph{arXiv preprint arXiv:1706.01307}, 2017.

\bibitem[Hendrycks and Gimpel(2016)]{gelu}
Dan Hendrycks and Kevin Gimpel.
\newblock Gaussian error linear units (gelus).
\newblock \emph{arXiv preprint arXiv:1606.08415}, 2016.

\bibitem[Howard and Gugger(2020)]{imagenette}
Jeremy Howard and Sylvain Gugger.
\newblock Fastai: a layered api for deep learning.
\newblock \emph{Information}, 11\penalty0 (2):\penalty0 108, 2020.

\bibitem[Kalogeropoulos et~al.(2024)Kalogeropoulos, Bouritsas, and Panagakis]{scale_equivariant_graph_metanetworks}
Ioannis Kalogeropoulos, Giorgos Bouritsas, and Yannis Panagakis.
\newblock Scale equivariant graph metanetworks.
\newblock \emph{arXiv preprint arXiv:2406.10685}, 2024.

\bibitem[Kingma(2014)]{adam}
Diederik~P Kingma.
\newblock Adam: A method for stochastic optimization.
\newblock \emph{arXiv preprint arXiv:1412.6980}, 2014.

\bibitem[Kofinas et~al.(2024)Kofinas, Knyazev, Zhang, Chen, Burghouts, Gavves, Snoek, and Zhang]{graph_equivariant_representations}
Miltiadis Kofinas, Boris Knyazev, Yan Zhang, Yunlu Chen, Gertjan~J Burghouts, Efstratios Gavves, Cees~GM Snoek, and David~W Zhang.
\newblock Graph neural networks for learning equivariant representations of neural networks.
\newblock \emph{arXiv preprint arXiv:2403.12143}, 2024.

\bibitem[Krizhevsky et~al.(2009)Krizhevsky, Hinton, et~al.]{cifar10}
Alex Krizhevsky, Geoffrey Hinton, et~al.
\newblock Learning multiple layers of features from tiny images.
\newblock 2009.

\bibitem[LeCun et~al.(1998)LeCun, Bottou, Bengio, and Haffner]{mnist}
Yann LeCun, L{\'e}on Bottou, Yoshua Bengio, and Patrick Haffner.
\newblock Gradient-based learning applied to document recognition.
\newblock \emph{Proceedings of the IEEE}, 86\penalty0 (11):\penalty0 2278--2324, 1998.

\bibitem[Li et~al.(2017)Li, Zhou, Chen, and Li]{meta_sgd}
Zhenguo Li, Fengwei Zhou, Fei Chen, and Hang Li.
\newblock Meta-sgd: Learning to learn quickly for few-shot learning.
\newblock \emph{arXiv preprint arXiv:1707.09835}, 2017.

\bibitem[Lim et~al.(2023)Lim, Maron, Law, Lorraine, and Lucas]{graph_processing_diverse}
Derek Lim, Haggai Maron, Marc~T Law, Jonathan Lorraine, and James Lucas.
\newblock Graph metanetworks for processing diverse neural architectures.
\newblock \emph{arXiv preprint arXiv:2312.04501}, 2023.

\bibitem[Loshchilov(2017)]{adamw}
I Loshchilov.
\newblock Decoupled weight decay regularization.
\newblock \emph{arXiv preprint arXiv:1711.05101}, 2017.

\bibitem[Luijmes et~al.(2025)Luijmes, Gielisse, Knyazhitskiy, and van Gemert]{luijmes2025arcanchoredrepresentationclouds}
Joost Luijmes, Alexander Gielisse, Roman Knyazhitskiy, and Jan van Gemert.
\newblock {ARC}: Anchored representation clouds for high-resolution {INR} classification, 2025.

\bibitem[Martel et~al.(2021)Martel, Lindell, Lin, Chan, Monteiro, and Wetzstein]{spatial_acorn}
Julien~NP Martel, David~B Lindell, Connor~Z Lin, Eric~R Chan, Marco Monteiro, and Gordon Wetzstein.
\newblock Acorn: Adaptive coordinate networks for neural scene representation.
\newblock \emph{arXiv preprint arXiv:2105.02788}, 2021.

\bibitem[Mildenhall et~al.(2021)Mildenhall, Srinivasan, Tancik, Barron, Ramamoorthi, and Ng]{nerf}
Ben Mildenhall, Pratul~P Srinivasan, Matthew Tancik, Jonathan~T Barron, Ravi Ramamoorthi, and Ren Ng.
\newblock Nerf: Representing scenes as neural radiance fields for view synthesis.
\newblock \emph{Communications of the ACM}, 65\penalty0 (1):\penalty0 99--106, 2021.

\bibitem[M{\"u}ller et~al.(2022)M{\"u}ller, Evans, Schied, and Keller]{spatial_instant_ngp}
Thomas M{\"u}ller, Alex Evans, Christoph Schied, and Alexander Keller.
\newblock Instant neural graphics primitives with a multiresolution hash encoding.
\newblock \emph{ACM transactions on graphics (TOG)}, 41\penalty0 (4):\penalty0 1--15, 2022.

\bibitem[Navon et~al.(2023{\natexlab{a}})Navon, Shamsian, Achituve, Fetaya, Chechik, and Maron]{dws_net}
Aviv Navon, Aviv Shamsian, Idan Achituve, Ethan Fetaya, Gal Chechik, and Haggai Maron.
\newblock Equivariant architectures for learning in deep weight spaces.
\newblock In \emph{International Conference on Machine Learning}, pages 25790--25816. PMLR, 2023{\natexlab{a}}.

\bibitem[Navon et~al.(2023{\natexlab{b}})Navon, Shamsian, Fetaya, Chechik, Dym, and Maron]{deep_weight_space_alignment}
Aviv Navon, Aviv Shamsian, Ethan Fetaya, Gal Chechik, Nadav Dym, and Haggai Maron.
\newblock Equivariant deep weight space alignment.
\newblock \emph{arXiv preprint arXiv:2310.13397}, 2023{\natexlab{b}}.

\bibitem[Papa et~al.(2024)Papa, Valperga, Knigge, Kofinas, Lippe, Sonke, and Gavves]{fit_a_nef}
Samuele Papa, Riccardo Valperga, David Knigge, Miltiadis Kofinas, Phillip Lippe, Jan-Jakob Sonke, and Efstratios Gavves.
\newblock How to train neural field representations: A comprehensive study and benchmark.
\newblock In \emph{Proceedings of the IEEE/CVF Conference on Computer Vision and Pattern Recognition}, pages 22616--22625, 2024.

\bibitem[Qi et~al.(2017{\natexlab{a}})Qi, Su, Mo, and Guibas]{pointnet}
Charles~R Qi, Hao Su, Kaichun Mo, and Leonidas~J Guibas.
\newblock Pointnet: Deep learning on point sets for 3d classification and segmentation.
\newblock In \emph{Proceedings of the IEEE conference on computer vision and pattern recognition}, pages 652--660, 2017{\natexlab{a}}.

\bibitem[Qi et~al.(2017{\natexlab{b}})Qi, Yi, Su, and Guibas]{pointnet_pp}
Charles~Ruizhongtai Qi, Li Yi, Hao Su, and Leonidas~J Guibas.
\newblock Pointnet++: Deep hierarchical feature learning on point sets in a metric space.
\newblock \emph{Advances in neural information processing systems}, 30, 2017{\natexlab{b}}.

\bibitem[Ramasinghe and Lucey(2022)]{gaussian_act}
Sameera Ramasinghe and Simon Lucey.
\newblock Beyond periodicity: Towards a unifying framework for activations in coordinate-mlps.
\newblock In \emph{European Conference on Computer Vision}, pages 142--158. Springer, 2022.

\bibitem[Saragadam et~al.(2022)Saragadam, Tan, Balakrishnan, Baraniuk, and Veeraraghavan]{spatial_miner}
Vishwanath Saragadam, Jasper Tan, Guha Balakrishnan, Richard~G Baraniuk, and Ashok Veeraraghavan.
\newblock Miner: Multiscale implicit neural representation.
\newblock In \emph{European Conference on Computer Vision}, pages 318--333. Springer, 2022.

\bibitem[Saragadam et~al.(2023)Saragadam, LeJeune, Tan, Balakrishnan, Veeraraghavan, and Baraniuk]{wire}
Vishwanath Saragadam, Daniel LeJeune, Jasper Tan, Guha Balakrishnan, Ashok Veeraraghavan, and Richard~G Baraniuk.
\newblock Wire: Wavelet implicit neural representations.
\newblock In \emph{Proceedings of the IEEE/CVF Conference on Computer Vision and Pattern Recognition}, pages 18507--18516, 2023.

\bibitem[Shamsian et~al.(2024)Shamsian, Navon, Zhang, Zhang, Fetaya, Chechik, and Maron]{weight_aug}
Aviv Shamsian, Aviv Navon, David~W Zhang, Yan Zhang, Ethan Fetaya, Gal Chechik, and Haggai Maron.
\newblock Improved generalization of weight space networks via augmentations.
\newblock \emph{arXiv preprint arXiv:2402.04081}, 2024.

\bibitem[Sitzmann et~al.(2020)Sitzmann, Martel, Bergman, Lindell, and Wetzstein]{siren}
Vincent Sitzmann, Julien Martel, Alexander Bergman, David Lindell, and Gordon Wetzstein.
\newblock Implicit neural representations with periodic activation functions.
\newblock \emph{Advances in neural information processing systems}, 33:\penalty0 7462--7473, 2020.

\bibitem[Tancik et~al.(2020)Tancik, Srinivasan, Mildenhall, Fridovich-Keil, Raghavan, Singhal, Ramamoorthi, Barron, and Ng]{fourier_features}
Matthew Tancik, Pratul Srinivasan, Ben Mildenhall, Sara Fridovich-Keil, Nithin Raghavan, Utkarsh Singhal, Ravi Ramamoorthi, Jonathan Barron, and Ren Ng.
\newblock Fourier features let networks learn high frequency functions in low dimensional domains.
\newblock \emph{Advances in neural information processing systems}, 33:\penalty0 7537--7547, 2020.

\bibitem[Tancik et~al.(2021)Tancik, Mildenhall, Wang, Schmidt, Srinivasan, Barron, and Ng]{meta_nerf}
Matthew Tancik, Ben Mildenhall, Terrance Wang, Divi Schmidt, Pratul~P Srinivasan, Jonathan~T Barron, and Ren Ng.
\newblock Learned initializations for optimizing coordinate-based neural representations.
\newblock In \emph{Proceedings of the IEEE/CVF Conference on Computer Vision and Pattern Recognition}, pages 2846--2855, 2021.

\bibitem[Touvron et~al.(2021)Touvron, Cord, Sablayrolles, Synnaeve, and J{\'e}gou]{layerscale}
Hugo Touvron, Matthieu Cord, Alexandre Sablayrolles, Gabriel Synnaeve, and Herv{\'e} J{\'e}gou.
\newblock Going deeper with image transformers.
\newblock In \emph{Proceedings of the IEEE/CVF international conference on computer vision}, pages 32--42, 2021.

\bibitem[Ulyanov et~al.(2018)Ulyanov, Vedaldi, and Lempitsky]{deep_image_prior}
Dmitry Ulyanov, Andrea Vedaldi, and Victor Lempitsky.
\newblock Deep image prior.
\newblock In \emph{Proceedings of the IEEE conference on computer vision and pattern recognition}, pages 9446--9454, 2018.

\bibitem[Vaswani(2017)]{attention_all_you_need}
A Vaswani.
\newblock Attention is all you need.
\newblock \emph{Advances in Neural Information Processing Systems}, 2017.

\bibitem[Wu et~al.(2015)Wu, Song, Khosla, Yu, Zhang, Tang, and Xiao]{modelnet40}
Zhirong Wu, Shuran Song, Aditya Khosla, Fisher Yu, Linguang Zhang, Xiaoou Tang, and Jianxiong Xiao.
\newblock 3d shapenets: A deep representation for volumetric shapes.
\newblock In \emph{Proceedings of the IEEE conference on computer vision and pattern recognition}, pages 1912--1920, 2015.

\bibitem[Xiao et~al.(2017)Xiao, Rasul, and Vollgraf]{fashionmnist}
Han Xiao, Kashif Rasul, and Roland Vollgraf.
\newblock Fashion-mnist: a novel image dataset for benchmarking machine learning algorithms.
\newblock \emph{arXiv preprint arXiv:1708.07747}, 2017.

\bibitem[Xu et~al.(2022)Xu, Wang, Jiang, Fan, and Wang]{inspnet}
Dejia Xu, Peihao Wang, Yifan Jiang, Zhiwen Fan, and Zhangyang Wang.
\newblock Signal processing for implicit neural representations.
\newblock \emph{Advances in Neural Information Processing Systems}, 35:\penalty0 13404--13418, 2022.

\bibitem[Zhou et~al.(2024{\natexlab{a}})Zhou, Yang, Burns, Cardace, Jiang, Sokota, Kolter, and Finn]{perm_eq_neural_functionals}
Allan Zhou, Kaien Yang, Kaylee Burns, Adriano Cardace, Yiding Jiang, Samuel Sokota, J~Zico Kolter, and Chelsea Finn.
\newblock Permutation equivariant neural functionals.
\newblock \emph{Advances in neural information processing systems}, 36, 2024{\natexlab{a}}.

\bibitem[Zhou et~al.(2024{\natexlab{b}})Zhou, Yang, Jiang, Burns, Xu, Sokota, Kolter, and Finn]{neural_functional_transformers}
Allan Zhou, Kaien Yang, Yiding Jiang, Kaylee Burns, Winnie Xu, Samuel Sokota, J~Zico Kolter, and Chelsea Finn.
\newblock Neural functional transformers.
\newblock \emph{Advances in neural information processing systems}, 36, 2024{\natexlab{b}}.

\end{thebibliography}
}

\newpage

\section{Supplementary Material}

\begin{table}[htbp]
\centering
\resizebox{\columnwidth}{!}{
\begin{tabular}{p{3cm} cccccccc}
\toprule

\textbf{Ablation} & \textbf{Parameter} & \textbf{Accuracy} & \textbf{PSNR} & \textbf{Validation} & \textbf{Training} & \textbf{Mem} & \textbf{\#Params} & \textbf{\#Params} \\
    & & (\%) & (dB) & (sec) & (min/epoch) & (GiB) & CLS & SIREN \\

\midrule

\multirow{6}{*}{\textbf{Inner-Loop Steps $k$}}
 & 1 & 47.38 & 	25.35 & 	6.7 & 	1.5 & 	2.2 & 	1.1M & 	133K \\
 & 2 & 45.15 & 	27.99 & 	8.2 & 	1.8 & 	2.4 & 	1.1M & 	199K \\
 & 4 & 43.12 & 	40.16 & 	11.0 & 	2.4 & 	2.7 & 	1.1M & 	332K \\
 & \underline{6} & 42.24 & 	43.11 & 	14.2 & 	3.1 & 	3.0 & 	1.1M & 	465K \\
 & 8 & 41.26 & 	43.55 & 	16.9 & 	3.7 & 	3.3 & 	1.1M & 	598K \\
 & 10 & 38.41 & 	38.46 & 	19.8 & 	4.3 & 	3.6 & 	1.1M & 	731K \\
\midrule
\multirow{6}{*}{\textbf{Scaling Factor $\lambda$}}
 & 1 & 40.53 & 	43.09 & 	13.9 & 	3.1 & 	3.0 & 	1.1M & 	465K \\
 & 10 & 43.60 & 	43.08 & 	13.9 & 	2.9 & 	3.0 & 	1.1M & 	465K \\
 & 100 & 42.67 & 	43.11 & 	13.9 & 	3.0 & 	3.0 & 	1.1M & 	465K \\
 & \underline{500} & 42.24 & 	43.11 & 	14.2 & 	3.1 & 	3.0 & 	1.1M & 	465K \\
 & 1000 & 42.28 & 	43.11 & 	14.1 & 	3.1 & 	3.0 & 	1.1M & 	465K \\
 & LayerNorm & 43.35 & 	43.11 & 	14.0 & 	3.0 & 	3.0 & 	1.1M & 	465K \\
\midrule
\multirow{4}{*}{\textbf{SIREN Depth}}
 & 2 & 41.76 & 	41.31 & 	8.9 & 	2.1 & 	1.5 & 	1.0M & 	234K \\
 & \underline{4} & 42.24 & 	43.11 & 	14.2 & 	3.1 & 	3.0 & 	1.1M & 	465K \\
 & 6 & 43.40 & 	45.61 & 	21.9 & 	4.5 & 	4.9 & 	1.1M & 	696K \\
 & 8 & 44.31 & 	48.56 & 	31.5 & 	6.4 & 	7.4 & 	1.1M & 	927K \\
\midrule
\multirow{3}{*}{\textbf{SIREN Width}}
 & 64 & 34.57 & 	34.45 & 	10.8 & 	2.8 & 	1.1 & 	273K & 	118K \\
 & \underline{128} & 42.24 & 	43.11 & 	14.2 & 	3.1 & 	3.0 & 	1.1M & 	465K \\
 & 256 & 46.45 & 	47.71 & 	45.0 & 	8.9 & 	12.7 & 	4.3M & 	1.8M \\
\midrule
\multirow{4}{*}{\textbf{Transformer Depth}}
 & 5 & 40.58 & 	43.14 & 	12.7 & 	2.6 & 	2.2 & 	581K & 	465K \\
 & \underline{10} & 42.24 & 	43.11 & 	14.2 & 	3.1 & 	3.0 & 	1.1M & 	465K \\
 & 15 & 43.45 & 	42.98 & 	16.1 & 	3.7 & 	3.7 & 	1.6M & 	465K \\
 & 20 & 44.70 & 	43.34 & 	17.6 & 	4.0 & 	4.5 & 	2.1M & 	465K \\
\midrule
\multirow{2}{*}{\textbf{Meta-SGD Shared}}
 & \underline{False} & 42.24 & 	43.11 & 	14.2 & 	3.1 & 	3.0 & 	1.1M & 	465K \\
 & True & 45.44 & 	39.12 & 	14.4 & 	3.2 & 	3.0 & 	1.1M & 	133K \\
\midrule
\multirow{3}{*}{\textbf{Meta-SGD LR}}
 & \underline{0.01} & 42.24 & 	43.11 & 	14.2 & 	3.1 & 	3.0 & 	1.1M & 	465K \\
 & 0.001 & 47.44 & 	39.17 & 	14.2 & 	3.0 & 	3.0 & 	1.1M & 	465K \\
 & 0.0001 & 48.71 & 	28.57 & 	14.2 & 	3.0 & 	3.0 & 	1.1M & 	465K \\
\midrule
\multirow{4}{*}{\textbf{Subsampling Rate $s$}}
 & \underline{1.0} & 42.24 & 	43.11 & 	14.2 & 	3.1 & 	3.0 & 	1.1M & 	465K \\
 & 0.1 & 47.65 & 	24.33 & 	11.9 & 	2.7 & 	2.1 & 	1.1M & 	465K \\
 & 0.01 & 43.34 & 	18.83 & 	12.2 & 	2.7 & 	2.0 & 	1.1M & 	465K \\
 & 0.001 & 25.35 & 	16.72 & 	12.7 & 	2.7 & 	2.0 & 	1.1M & 	465K \\

\bottomrule
\end{tabular}
}

\caption{Ablation results for the proposed WT model, which is MWT with $w_{task}$ set to 0. This model trained on 80\% of the CIFAR-10 training dataset. The reported numbers are evaluated on the remaining 20\% of the training set that was held out for validation. Accuracy indicates the percentage of correctly classified samples, while PSNR represents the average reconstruction quality of the image. Based on the table, we also compute a variant of WT with $s=0.1$, Meta-SGD learning rate of $0.0001$, and stepwise shared Meta-SGD. This model achieves an accuracy of $49.8\%$ and PSNR of $23.1$ dB, which still performs worse than our MWT model.}
\label{table:ablation_wt}
\end{table}

\begin{table}[htbp]
\centering
\resizebox{\columnwidth}{!}{
\begin{tabular}{p{3cm} cccccccc}
\toprule

\textbf{Ablation} & \textbf{Parameter} & \textbf{Accuracy} & \textbf{PSNR} & \textbf{Validation} & \textbf{Training} & \textbf{Mem} & \textbf{\#Params} & \textbf{\#Params} \\
    & & (\%) & (dB) & (sec/epoch) & (min/epoch) & (GiB) & CLS & SIREN \\

\midrule
\multirow{6}{*}{\textbf{Inner-Loop Steps $k$}}
 & 1 & 48.10 & 	25.35 & 	6.8 & 	1.5 & 	2.2 & 	1.1M & 	133K \\
 & 2 & 46.15 & 	27.99 & 	8.2 & 	1.8 & 	2.4 & 	1.1M & 	199K \\
 & 4 & 40.90 & 	40.17 & 	11.1 & 	2.4 & 	2.6 & 	1.1M & 	332K \\
 & \underline{6} & 41.85 & 	43.12 & 	13.9 & 	3.0 & 	3.0 & 	1.1M & 	465K \\
 & 8 & 42.72 & 	43.53 & 	16.8 & 	3.5 & 	3.3 & 	1.1M & 	598K \\
 & 10 & 42.42 & 	38.46 & 	21.5 & 	4.4 & 	3.6 & 	1.1M & 	731K \\
\midrule
\multirow{6}{*}{\textbf{Scaling Factor $\lambda$}}
 & 1 & 40.52 & 	43.12 & 	13.9 & 	3.0 & 	3.0 & 	1.1M & 	465K \\
 & 10 & 42.45 & 	43.12 & 	14.1 & 	3.0 & 	3.0 & 	1.1M & 	465K \\
 & 100 & 41.59 & 	43.12 & 	14.4 & 	3.1 & 	3.0 & 	1.1M & 	465K \\
 & \underline{500} & 41.85 & 	43.12 & 	13.9 & 	3.0 & 	3.0 & 	1.1M & 	465K \\
 & 1000 & 42.14 & 	43.12 & 	14.0 & 	3.0 & 	3.0 & 	1.1M & 	465K \\
 & LayerNorm & 42.95 & 	43.12 & 	14.0 & 	3.0 & 	3.0 & 	1.1M & 	465K \\
\midrule
\multirow{4}{*}{\textbf{SIREN Depth}}
 & 2 & 43.14 & 	41.36 & 	8.8 & 	2.1 & 	1.5 & 	1.0M & 	234K \\
 & \underline{4} & 41.85 & 	43.12 & 	13.9 & 	3.0 & 	3.0 & 	1.1M & 	465K \\
 & 6 & 44.89 & 	45.62 & 	21.9 & 	4.5 & 	4.9 & 	1.1M & 	696K \\
 & 8 & 44.97 & 	48.68 & 	31.4 & 	6.3 & 	7.4 & 	1.1M & 	927K \\
\midrule
\multirow{3}{*}{\textbf{SIREN Width}}
 & 64 & 38.01 & 	34.46 & 	9.5 & 	2.7 & 	1.1 & 	273K & 	118K \\
 & \underline{128} & 41.85 & 	43.12 & 	13.9 & 	3.0 & 	3.0 & 	1.1M & 	465K \\
 & 256 & 49.39 & 	47.71 & 	43.6 & 	8.7 & 	12.7 & 	4.3M & 	1.8M \\
\midrule
\multirow{4}{*}{\textbf{Transformer Depth}}
 & 5 & 39.22 & 	43.14 & 	12.3 & 	2.6 & 	2.2 & 	581K & 	465K \\
 & \underline{10} & 41.85 & 	43.12 & 	13.9 & 	3.0 & 	3.0 & 	1.1M & 	465K \\
 & 15 & 41.82 & 	42.99 & 	15.8 & 	3.5 & 	3.7 & 	1.6M & 	465K \\
 & 20 & 46.44 & 	43.34 & 	18.6 & 	4.0 & 	4.5 & 	2.1M & 	465K \\
\midrule
\multirow{2}{*}{\textbf{Meta-SGD Shared}}
 & \underline{False} & 41.85 & 	43.12 & 	13.9 & 	3.0 & 	3.0 & 	1.1M & 	465K \\
 & True & 44.91 & 	39.13 & 	13.9 & 	3.0 & 	3.0 & 	1.1M & 	133K \\
\midrule
\multirow{3}{*}{\textbf{Meta-SGD LR}}
 & \underline{0.01} & 41.85 & 	43.12 & 	13.9 & 	3.0 & 	3.0 & 	1.1M & 	465K \\
 & 0.001 & 48.93 & 	39.17 & 	14.0 & 	3.1 & 	3.0 & 	1.1M & 	465K \\
 & 0.0001 & 49.82 & 	28.57 & 	14.0 & 	3.0 & 	3.0 & 	1.1M & 	465K \\
\midrule
\multirow{4}{*}{\textbf{Subsampling Rate $s$}}
 & \underline{1.0} & 41.85 & 	43.12 & 	13.9 & 	3.0 & 	3.0 & 	1.1M & 	465K \\
 & 0.1 & 49.27 & 	24.32 & 	11.9 & 	2.7 & 	2.1 & 	1.1M & 	465K \\
 & 0.01 & 45.73 & 	18.74 & 	11.9 & 	2.6 & 	2.0 & 	1.1M & 	465K \\
 & 0.001 & 25.83 & 	16.63 & 	11.9 & 	2.8 & 	2.0 & 	1.1M & 	465K \\

\bottomrule
\end{tabular}
}

\caption{Ablation results for a sequential version of the WT model. Here, we use the WT model, but we do not train the meta-learning of the INR simultaneously with training the classifier, as is the case for WT and MWT. Instead, we first train the meta-learning INR to accurately reconstruct, and then train a classifier on these representations afterwards, as this two-step approach is commonly done in existing work. We do not use this model in the paper itself due to high similarity in results with the WT model. This model trained on 80\% of the CIFAR-10 training dataset. The reported numbers are evaluated on the remaining 20\% of the training set that was held out for validation. Accuracy indicates the percentage of correctly classified samples, while PSNR represents the average reconstruction quality of the image.}
\label{table:ablation_wt_seq}
\end{table}

\begin{table}[H]
\centering
\resizebox{\columnwidth}{!}{%
\begin{tabular}{@{}lccccc@{}}
\toprule
\textbf{MWT CIFAR-10} & \multicolumn{5}{c}{\textbf{Different number of steps at test time}} \\
(6 train steps) & 6 steps & 8 steps & 10 steps & 15 steps & 20 steps \\
\midrule
\textbf{Accuracy}       &  \textbf{55.5\%} & 55.1\% & 54.9\% & 53.5\% & 53.2\%  \\
\textbf{PSNR}           &  30.5 dB & 31.6 dB & 32.9 dB & 33.9 dB & \textbf{35.2 dB} \\
\bottomrule
\end{tabular}%
}
\caption{We increase the amount of update steps at test time, for a model trained with $k=6$. Note that this requires sharing the Meta-SGD learning rates over the update steps. We show that this increases reconstruction quality, but at cost of classification performance, similar to the observations by \cite{fit_a_nef}. }
\label{tab:subsampling}
\end{table}

\begin{figure}
    \centering
    \includegraphics[width=0.8\linewidth]{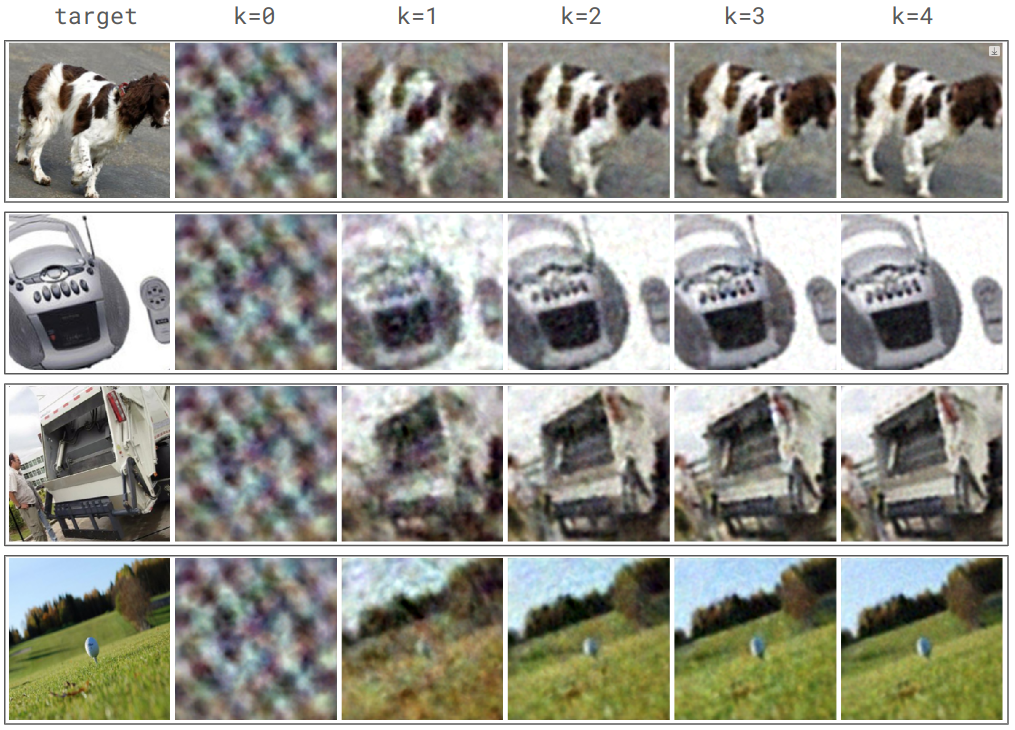}
    \caption{Our proposed meta-learning framework can quickly adapt to a new signal. This figure demonstrates how a SIREN converges on high-resolution Imagenette images in just $k=4$ update steps, while also providing INR weights that are structured for classification. The images shown here come from $\text{MWT-L}_{s=0.1}$ trained on Imagenette, and have an average PSNR of $22.31$ dB after the last step $k=4$.}
    \label{fig:reconstruct_preview}
\end{figure}

\end{document}